\documentclass[
]{ceurart}

\usepackage{listings}
\usepackage{float}
\usepackage{wrapfig}
\usepackage{booktabs}
\usepackage{multirow}
\usepackage{makecell}
\usepackage{xcolor}
\usepackage{array}
\usepackage{caption}
\usepackage{rotating}

\usepackage{siunitx}

\begin{document}

\copyrightyear{2026}
\copyrightclause{Copyright for this paper by its authors.
  Use permitted under Creative Commons License Attribution 4.0
  International (CC BY 4.0).}

\conference{CLiC-it 2026: Twelfth Italian Conference on Computational Linguistics, September 14 — 16, 2026, Palermo, Italy}

\title{"Don't Say It!": Constraints, Compliance, and Communication when Language Models Play Taboo}


\author[1]{Sara Candussio}[
orcid=0009-0004-5198-6970,
email=sara.candussio@phd.units.it,
]
\fnmark[1]
\address[1]{AILab, MIGe, Univeristy of Trieste, Italy}

\author[2]{Francesca Padovani}[
orcid=0009-0007-3489-9631,
email=f.padovani@rug.nl,
]
\fnmark[1]
\address[2]{Center for Language and Cognition (CLCG), University of Groningen, The Netherlands}

\author[2,3]{Daniel Scalena}[%
orcid=0009-0006-0518-6504,
email=d.scalena@rug.nl,
]
\fnmark[1]
\address[3]{University of Milano - Bicocca, Italy}

\author[2]{Malvina Nissim}[%
orcid=0000-0001-5289-0971,
email=m.nissim@rug.nl
]

\cortext[1]{Corresponding author.}
\fntext[1]{These authors contributed equally.}


\begin{abstract}
The game of Taboo requires describing a target word without using a set of forbidden words, so that other players can guess it. This deceptively simple task combines strict lexical constraints with the need for communicatively effective descriptions, making it a compelling playground for examining how LLMs navigate competing demands at inference time. We evaluate two open-weight models under conditions that intervene at progressively deeper levels of the generative process, from prompting to generation-time constraints to internal representations manipulations. We assess their outputs through forbidden word violation detection, LLM-as-a-judge measuring the degree to which generated descriptions successfully evoke the target concept for both human and machine guessers, and examining whether the strategies models adopt under constraint align with those of human players. Our results show that compliance with the rules of the game and communicative effectiveness trade off differently across conditions, and that models remain substantially weaker than humans as guessers, suggesting that lexical grounding under constraint is an open challenge for current language models\footnote{Code and data can be found at \url{https://github.com/DanielSc4/LMtaboo/}}.
\end{abstract}


\maketitle

\section{Introduction}

The game of Taboo requires a player to describe a target word without using a set of forbidden words, so that another player can guess it. Beyond its appeal as a \textit{parlour} game, Taboo instantiates a linguistically interesting challenge: the speaker must suppress some lexically and conceptually salient words while simultaneously producing a description that is informative enough to identify the target. This combination of constraint satisfaction and communicative effectiveness makes it a particularly suitable setting for probing language models in a setting that goes beyond standard instruction-following benchmarks. 
In particular, it remains unclear whether descriptions generated under constraint conform to the demands of the task --- namely whether they are sufficiently detailed, salient enough to evoke the target concept, and communicatively effective enough for the game to be successfully completed.

In this paper we investigate how LLMs play Taboo in Italian, evaluating two open-weight models under conditions that intervene at progressively deeper levels of the generative process: from prompting, to generation-time constraints, to the manipulation of internal representations. To ground our evaluation in an actual game-play, we conduct a human study in which the roles are reversed in both directions: human evaluators read model-generated descriptions and attempt to guess the target word, and vice versa. This bidirectional design allows us to directly assess how well descriptions produced succeed in conveying the target concept, and whether the descriptive choices adopted by models and humans are mutually interpretable.


Our findings reveal that the relationship between rule compliance and description quality varies substantially across conditions, and that models fall considerably short of human performance in the guessing role, pointing to lexical grounding under constraint as an open challenge for current language models.

\section{Related works}
\label{sec:related}

\paragraph{Language games as NLP benchmarks}
Word-based games have been exploited as playgrounds for NLP, offering well-defined rules and measurable objectives suitable for benchmarking. Within the Italian NLP community, language games have attracted growing attention as probes for linguistic competence: the TV game \textit{La Ghigliottina} has been tackled both with dedicated NLP systems \citep{sangati-etal-2020-challenge} and used to benchmark LLMs \citep{manna-etal-2024-riddle}, while studies on rebus and crossword solving reveal consistent limitations in tasks requiring compositional, phonological, and lateral reasoning \citep{sarti-etal-2024-non,sarti-etal-2024-eurekarebus, ciaccio-etal-2025-crossword,ciaccio2026cruciverbit}. Beyond Italian, benchmarks such as NYT Connections \citep{loredo-lopez-etal-2025-nyt} and Codenames \citep{stephenson2025codenamesbenchmarklargelanguage} consistently show a substantial gap between LLM and human performance on tasks requiring deliberate reasoning and theory of mind. Word games have also been studied as training environments: \citet{horst-etal-2025-playpen} show that LLMs can learn from dialogue game self-play feedback, with Taboo among the training games.  Closest to our work, \citet{cywinski2025elicitinglatentknowledgellms} study Taboo specifically in the context of LLM interpretability, using the game as a test for eliciting latent knowledge. Our work differs in scope: rather than focusing on a single intervention, we systematically compare constraint enforcement strategies at multiple levels of the generative process and we furthermore ground the evaluation in an actual human game-play.

\paragraph{Constrained text generation}
Comprehensive evaluations of constrained text generation across open-source LLMs show that models exhibit systematic deficiencies even on relatively simple lexical constraints. Prompt-based approaches suffer from position bias and struggle with morphologically complex forms, and stricter format constraints have been demonstrated to cause performance degradation \citep{yao2024collie,tam-etal-2024-speak}. \citet{ciaccio-etal-2024-controllable} evaluate several Italian LLMs on their ability to generate sentences adhering to morpho-syntactic specifications, finding systematic deficiencies across models and constraint types. \citet{calderaro-etal-2025-oulibench} further show that even state-of-the-art proprietary models consistently fall short when faced with fine-grained linguistic restrictions in Italian and that stricter requirements tend to degrade output quality more severely. These findings motivate our decision to go beyond prompting and to opt for different types of interventions.

\paragraph{Mechanistic interpretability and SAEs}
Beyond prompting, a more direct approach to output control is logit masking, i.e. forcing forbidden token probabilities to $-\infty$ at every decoding step, as implemented in frameworks such as that proposed by \citet{willard2023efficientguidedgenerationlarge} and its predecessors \citet{hokamp-liu-2017-lexically}. At a deeper level, concept manipulation via internal model representations offers an alternative approach. Rather than suppressing surface tokens, it targets the concepts encoded in the model's latent space using Sparse AutoEncoders (SAEs) \citep{cunningham2023sparseautoencodershighlyinterpretable}. They decompose dense LLM activations into sparse, monosemantic features \citep{templeton2024scaling} that can be ablated at inference time without modifying model weights. Furthermore, a feature's interpretability does not guarantee its effectiveness as a steering target: features that cleanly represent a concept may have little causal influence on the model's output -- a gap whose limits we help characterise in the context of lexical avoidance.

\begin{wrapfigure}{r}{0.4\textwidth}
  \centering
  \vspace{-2cm}
  \includegraphics[width=0.37\textwidth]{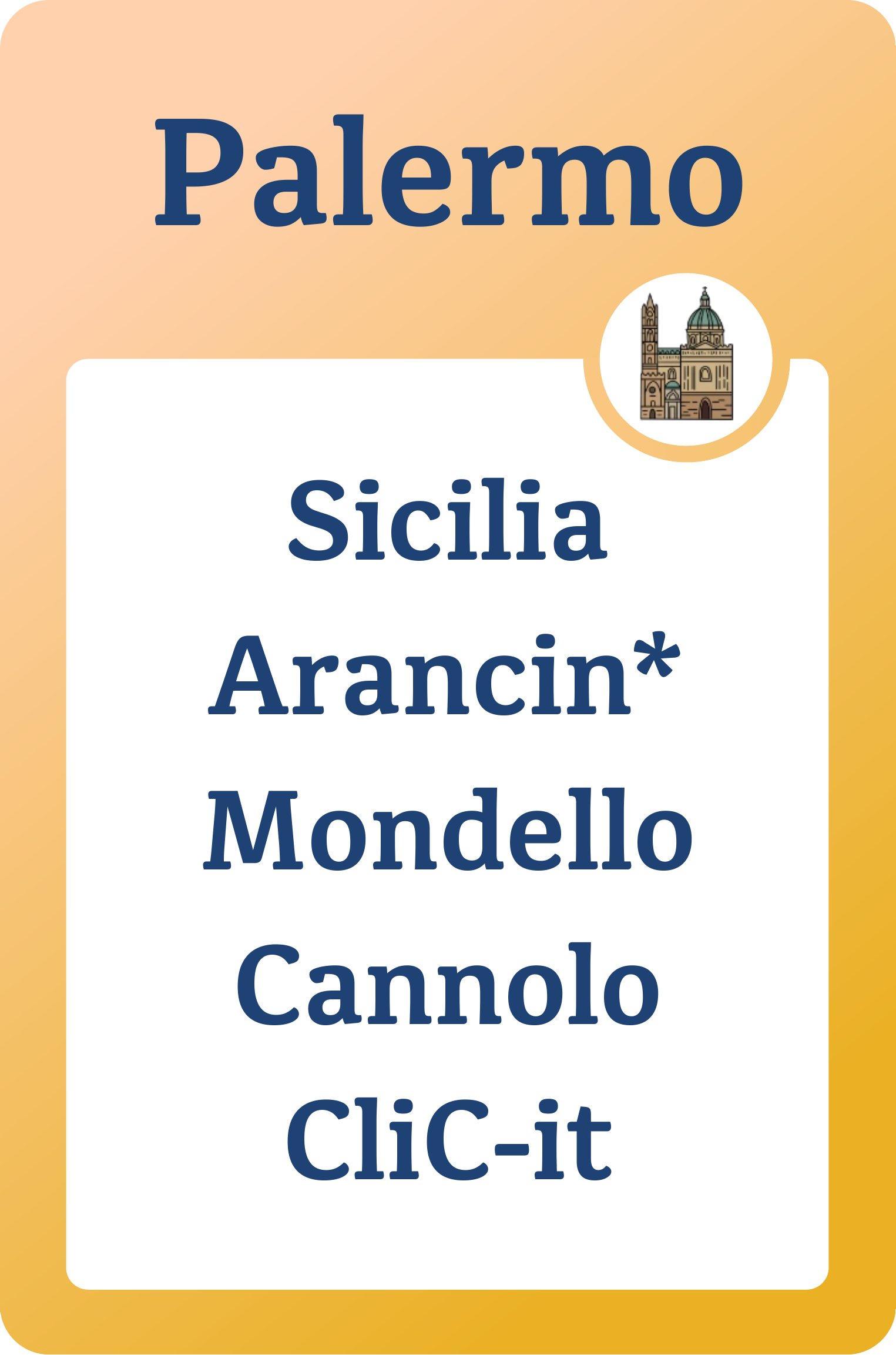}
  \caption{An example of a Taboo card. The  word to be guessed is on top. The five words  below are forbidden from appearing in the description.}
  \label{fig:carta}
\end{wrapfigure}

\section{Methodology}

\subsection{Data and models}

The datasets consists of 194 italian Taboo cards manually transcribed from a physical version of the board game. Each example contains a target word and a list of forbidden words that cannot be used in the description. 

We experiment with two open-weight models. \textbf{\texttt{google/gemma-3-4b-it}} \citep{gemma_2025} is a lightweight instruction-tuned model selected for its strong general performance and for the availability of pre-trained SAEs\footnote{\texttt{google/gemma-scope-2-4b-it} \cite{lieberum-etal-2024-gemma}} compatible with our interpretability-based experimental condition. \textbf{\texttt{openai/gpt-oss-20b}} \citep{openai2025gptoss120bgptoss20bmodel} is a Mixture-of-Experts reasoning model post-trained with chain-of-thought reinforcement learning, chosen to assess the role of explicit reasoning in constrained description generation, evaluated with and without reasoning. Generating an effective Taboo description requires implicitly planning ahead: the model must produce a description that is semantically informative while simultaneously avoiding a set of lexically related forbidden words. We therefore expect explicit reasoning to play a meaningful role in this task.

\subsection{Forbidding experiments}
We investigate three degrees of forbidding, each intervening at a different level of the model's behaviour. In the first condition, models are \emph{prompted} with the target word and the list of forbidden words, analogously to how human players are instructed before their turn. The second condition moves the intervention downstream, \emph{constraining the model during generation} so as to directly block forbidden tokens from being produced. The third condition operates at the level of internal representations, \emph{manipulating the model's latent space} to suppress the encoded concepts corresponding to the forbidden words.

\subsubsection{Prompting}
\label{sec:prompting}

The prompt is constructed programmatically for each card, embedding the list of forbidden words and the target word directly into the user message. To ensure clean and comparable outputs, we additionally provide output instructions specifying that the model should return only the description, with no surrounding text, and within a limit of 30 words. The prompt explicitly instructs the model to avoid not only the forbidden words verbatim, but also any word derived from the same morphological root, in order to fully comply with the rules of the game. The full prompt structure is provided in Table~\ref{tab:prompt_example} of Appendix~\ref{app:prompts}. Notably, the prompt does not explicitly instruct the model to avoid using the target word itself in the description, an error that would trivially make the model wrong at the game. We deemed it preferable to assess the extent to which this occurs empirically, and defer this analysis to Section~\ref{sec:evaluation}. In Appendix~\ref{app:prompts} we additionally report results obtained with a relaxed prompt variant that omits the morphological constraint, discussing the impact of this simplification on model behaviour.

\subsubsection{Generation-time constraints}

\paragraph{\textit{A priori} stems censorship}
Before generation begins, we compute a static set of banned token from the forbidden word list. For each forbidden word, we extract all stems and ban any vocabulary token whose surface form is a prefix of any such stem or lemma, or vice versa. The banned token are then masked by forcing their logits to $-\infty$ at every decoding step, making them impossible to sample regardless of context.
A key design choice is the use of morphological normalisation rather than exact string matching: the censorship extends beyond the literal forbidden words to their inflected and derived forms. As a consequence, the model is forced to use synonyms or periphrases to express the forbidden elements. Complete freedom is left to the reasoning process. 

\paragraph{Inference-time constrained generation}
A limitation of the \textit{a-priori} approach is that it cannot condition the ban on generation context, occasionally suppressing tokens that would be appropriate in the description but happen to share a morphological root with a forbidden word. A practical example of its aggressiveness can be found in Appendix \ref{app:a_priori}. We address this by constraining the reasoning process dynamically. Rather than banning tokens upfront, we allow the model to generate freely and intervene only when a complete word is detected; its stem is then compared against the forbidden ones. If a match is found, generation backtracks to the first token of the offending word, which is added to a persistent position-level mask, and generation resumes from that position with the triggering token suppressed. 
Compared to the \textit{a priori} approach, it operates on complete surface forms in place of subword fragments, avoiding spurious bans and higher computational overhead due to potential backtracing.

\subsubsection{Model Manipulations}

We also explore a representation-based approach that intervenes directly on the model's residual stream at generation time. The intervention operates in two stages: feature identification and steered generation.
\paragraph{Feature identification}
For each forbidden word $w_i$, we identify the SAE feature most strongly associated with it by running the model on the short natural prompt "\textit{La parola vietata è $w_i$}" (translated: "\textit{The forbidden word is $w_i$}") and extracting the residual-stream activations at layer 29 of \texttt{google/gemma-3-4b-it}\footnote{Layer 29 corresponds to $\sim85\%$ network depth, where representations are richer in semantic content (see Appendix~\ref{app:sae-impl}).}. These activations are passed through a pre-trained JumpReLU SAE from \texttt{google/gemma-scope-2-4b-it} \citep{lieberum-etal-2024-gemma}, using a residual-post hook at width 65k. The feature with the highest activation across the token span corresponding to $w_i$ is selected as its representative. This process is applied to all forbidden words in a card, yielding a set of feature indices $\{ f_1, \ldots, f_k \}$.

\paragraph{Steered generation} At generation time, we register a forward hook on the same residual-stream layer. At each decoding step, we compute the mean decoder vector $d=\frac{1}{k}\sum_j \mathbf{w}^{\text{dec}}_{f_i}$ across the selected features and apply an activation-norm-scaled intervention:

\[
\mathbf{h}' = \mathbf{h} + \alpha \cdot \|\mathbf{h}\|_2 \cdot \mathbf{d}
\]

steering the residual stream \emph{away} from the direction associated with the forbidden concepts.
The model is otherwise prompted without any mention of forbidden words, receiving only the target word and the standard output instructions. The entire process operates at inference time without modifying model weights.

\subsection{Evaluation}
\label{sec:evaluation}
\paragraph{Exact amount of violations}
We evaluate constraint compliance in the description generation through a number of metrics. 
\textit{Accuracy} measures the percentage of valid descriptions that fully comply with the rules of the game, i.e. outputs that contain neither verbatim forbidden words nor any morphologically related form. \textit{Share of Exact Violations} reports the percentage of descriptions in which the model uses one of the forbidden words verbatim. \textit{Share of Morphological Violations} reports the percentage of descriptions in which the model produces a word sharing the same stem or lemma as a forbidden word\footnote{Detected using the \texttt{it\_core\_news\_sm} spaCy model~\cite{honnibal2020spacy} and the \texttt{nltk} Italian Snowball stemmer~\cite{bird-loper-2004-nltk}.}. \textit{Share of no} \texttt{<eot>} applies exclusively to \texttt{openai/gpt-oss-20b} in reasoning mode and it measures the proportion of cases in which the model exhausts its 2048-token reasoning budget without producing a final description. Finally, as anticipated in Section~\ref{sec:prompting}, the prompt does not explicitly instruct the model to avoid using the target word itself in the description. \textit{Share of Target Leaked Words} reports the percentage of outputs in which the model nevertheless commits this error, inadvertently revealing the answer it was supposed to help guess.

\paragraph{LLM-as-a-judge}
To assess the communicative effectiveness of the generated descriptions, we employ an LLM-as-a-judge evaluation~\cite{bavaresco-etal-2025-llms} as a complement to human evaluation. We use both \texttt{google/gemma-3-4b-it} and \texttt{openai/gpt-oss-20b} as automatic judges, evaluating descriptions generated by each model with both judges --- that is, each model judges its own outputs as well as those of the other. Each judge is presented with the generated description alone, with no access to the forbidden words or the target word, and asked to guess the word being described. Instead of generating a free-form answer, we extract the judge's next-token probability distribution immediately after reading the description and measure how much probability mass is assigned to the first token of the target word. This allows for a fine-grained, continuous assessment of how strongly the description evokes the target concept, rather than a binary correct/incorrect judgment.

We derive two metrics from this distribution: \textit{Pass@$k$} measures whether the target word's first token falls within the top-$k$ most likely tokens, $k \in \{1, 2, 3, 5, 10\}$; the \textit{Raw token likelihood} reports the actual probability assigned to the target. 

\paragraph{Human Evaluation}
\label{sec:human_eval}
We conduct a human evaluation study involving 8 annotators, structured into two phases on disjoint sets of cards.

In the \textbf{guessing phase} (80 cards), annotators are presented with model-generated descriptions and asked to guess the target word as if playing an actual game of Taboo, albeit without the interactive component. Each annotator sees 20 cards, with system conditions distributed randomly across annotators to ensure balanced coverage of all evaluated systems, and each card evaluated under two different conditions by two different annotators. 

In the \textbf{description phase} (100 cards, disjoint from the guessing phase), each annotator receives 20 cards. For 5 of these, they produce a spontaneous, oral-style description; for the other 5, they write a deliberate, carefully worded description. This design mirrors the \emph{without thinking} and \emph{with thinking} regimes observed in model experiments respectively. These 10 descriptions of unique items produced by each annotator are then presented to the other annotators for guessing, with the oral description setting being the closest to the actual Taboo game. 
The last 10 cards are held constant across all annotators: every participant describes the same items, enabling future analysis of inter-annotator variability in description strategies, which we leave for future work.

\section{Results}

\subsection{Models are quite good at following given constraints}
\begin{figure*}[h]
    \centering
    \includegraphics[width=1\linewidth]{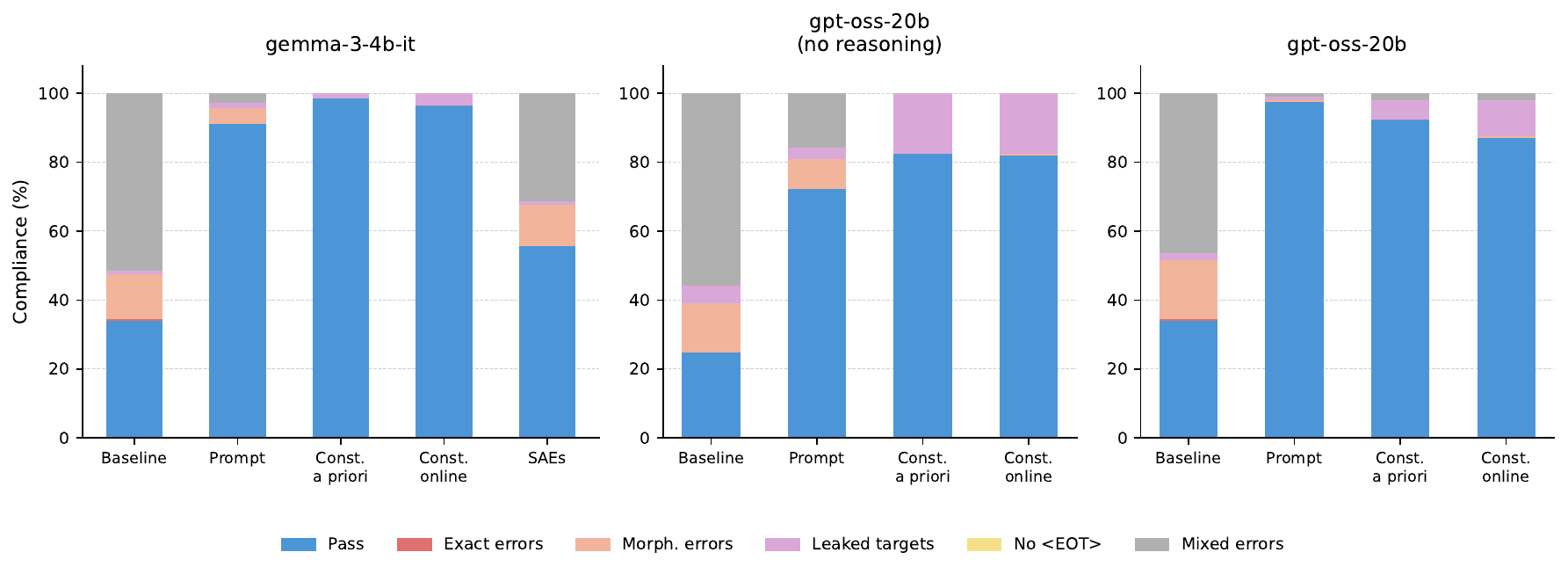}
    \caption{Compliance breakdown by method and model. Bars are partitioned into mutually exclusive categories summing to 100\%; outputs with multiple error types are grouped into Mixed. The Baseline frequently violates taboo rules, while prompting substantially reduces errors. Constrained decoding achieves near-perfect compliance by construction, with residual failures almost exclusively due to target-word leaks. For \texttt{gpt-oss-20b}, enabling reasoning further improves compliance.}
    \label{fig:static-eval}
\end{figure*}

Figure~\ref{fig:static-eval} provides an overview of model performance across forbidding experiments, compared against a \textit{baseline} in which models generate target word descriptions without any forbidden word constraint.\footnote{Prompt in Appendix~\ref{app:prompts}; full results in Tables~\ref{tab:results-gemma}--\ref{tab:results-gptoss-repen} in  Appendix~\ref{app:results}.} Baseline accuracy varies considerably across models, ranging from 24.7\% for \texttt{openai/gpt-oss-20b} without reasoning to 34.0\% for \texttt{google/gemma-3-4b-it}, reflecting how readily each model would spontaneously use the forbidden words for descriptive purposes, even without knowing they are banned.

Under the prompting constraint, accuracy rises markedly for both \texttt{google/gemma-3-4b-it} (91.2\%) and \texttt{openai/gpt-oss-20b} with reasoning (97.4\%). For \texttt{google/gemma-3-4b-it}, a small proportion of exact and morphological violations persist; \texttt{openai/gpt-oss-20b} with reasoning exhibits only morphological violations, suggesting that chain-of-thought is effective at suppressing verbatim forbidden words but less reliable on morphologically related forms. When reasoning is disabled, accuracy drops to 72.2\%, with a higher share of both violation types, confirming that the reasoning process plays an active role in enforcing lexical constraints.

In the constrained generation setting, all three models achieve high accuracy, with \texttt{google/gemma-3-4b-it} reaching near-perfect compliance under both a priori (98.5\%) and online (96.4\%) settings. \texttt{openai/gpt-oss-20b} with reasoning performs similarly under a priori constraints (92.3\%), though accuracy drops in the online condition (87.1\%), accompanied by a higher rate of target word leaks. Without reasoning, the model's accuracy reaches 82.5\% and 82.0\% respectively, with a markedly higher proportion of leaked target words in both cases. 


Finally, the SAE-based approach (Section~\ref{sec:related}), evaluated on \texttt{google/gemma-3-4b-it} only, reaches 55.7\%, well below both prompting (91.2\%) and constrained generation (98.5\% and 96.4\%), confirming that representation-based interventions alone cannot match explicit lexical guidance, though the result remains well above the unconstrained baseline (34.0\%).

\subsection{The type of constraining strategy shapes description informativeness}

\begin{figure*}[h]
    \centering
    \includegraphics[width=1\linewidth]{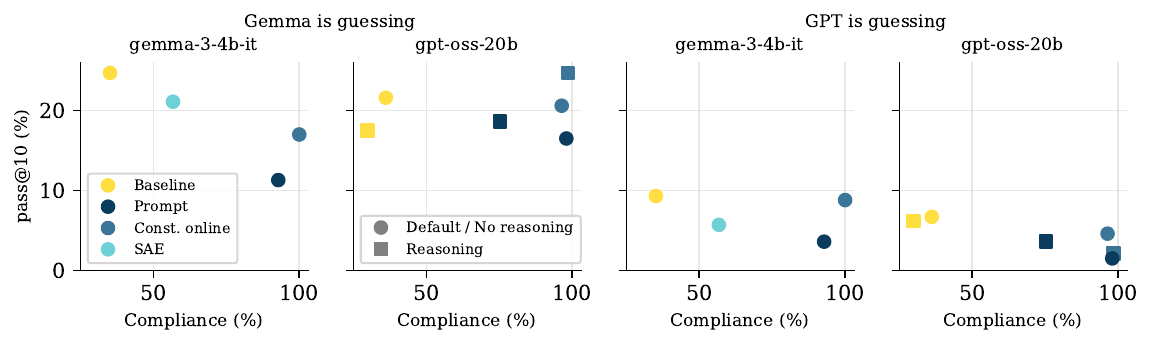}
    \caption{
        Compliance vs. pass@10 across models, methods, and evaluators. All three proposed methods (Prompt, Constrained, SAE) substantially increase compliance over the Baseline, confirming their effectiveness at enforcing the Taboo constraint. Pass@10 remains broadly stable across methods, suggesting compliance gains do not come at the cost of description quality. Gemma, in the guesser role, consistently achieves higher pass@10 than GPT both in- and out-of-distribution. Finally, enabling reasoning in gpt-oss-20b does not yield consistent improvements in its description effectiveness.
    }
    \label{fig:judge-eval}
\end{figure*}

Having established that constraining strategy shapes compliance, we now ask whether it also affects how informative the resulting descriptions are as measured by the ability of a judge model to identify the target word.

The two judge models agree on the broad ordering of conditions but differ substantially in sensitivity: \texttt{openai/gpt-oss-20b} assigns near-zero probability to the target word across almost all conditions, with pass@10 never exceeding 10\%; \texttt{google/gemma-3-4b-it} is consistently more generous, reaching up to 24.7\% on the same descriptions (Figure~\ref{fig:judge-eval}).

Despite this gap, both judges agree that the constraining strategy matters. Descriptions produced under inference-time constrained generation are in most conditions easier to guess than prompt-based ones: for \texttt{google/gemma-3-4b-it} as clue-giver, constrained online generation yields pass@10 of 17.0\% and 8.8\% against 11.3\% and 3.6\% for prompting under the \texttt{google/gemma-3-4b-it} and \texttt{openai/gpt-oss-20b} judge respectively. A likely explanation is that forcing the model away from forbidden tokens at decoding time pushes it toward less obvious but more semantically precise paraphrases, whereas an instructed model tends toward overly cautious, informationally sparse descriptions.

SAE-based steering presents the most striking dissociation: compliance is low (56.7\% strict accuracy) yet communicative effectiveness is competitive, with pass@10 reaching 21.1\% under the \texttt{google/gemma-3-4b-it} judge, also exceeding constrained generation for the same clue-giver. This suggests that blocking surface tokens at decoding time imposes a descriptive cost that representation-level intervention avoids: the model loses access to forbidden words but retains the underlying conceptual associations needed to produce an informative description.

\subsection{Models are weak guessers}
\label{sec:model_guess}

\begin{figure*}[h]
    \centering
    \includegraphics[width=\linewidth]{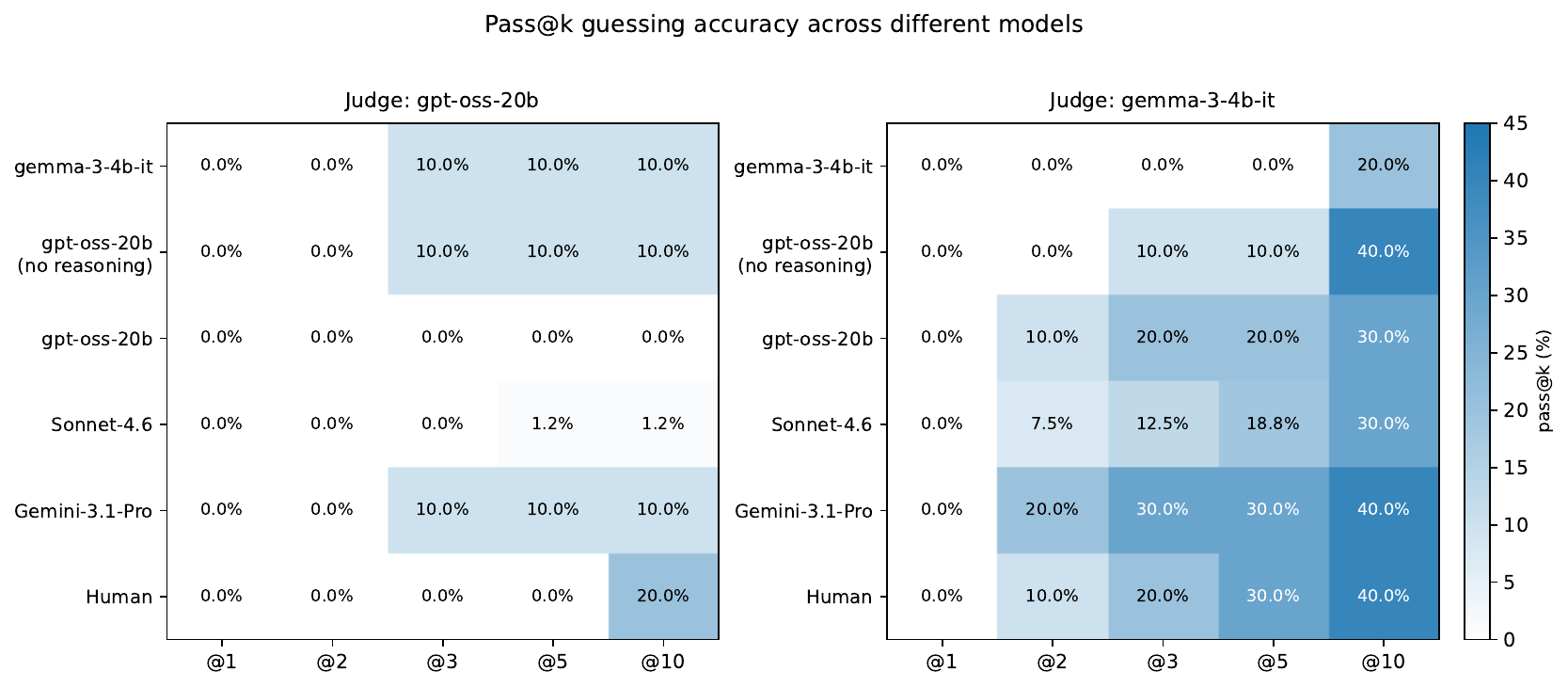}
        \caption{Pass@$k$ guessing accuracy on 10 shared Taboo cards, for all clue-givers evaluated by \texttt{openai/gpt-oss-20b} (left) and \texttt{google/gemma-3-4b-it} (right). Each cell reports the fraction of cards for which the judge ranked the target word within its top $k$ guesses.}
    \label{fig:models-vs-models}
\end{figure*}

We generated descriptions for 10 shared Taboo cards using \texttt{google/gemma-3-4b-it}, \texttt{openai/gpt-oss-20b} (with and without reasoning), Claude Sonnet-4.6 (max effort)~\cite{anthropic2026claudesonnet46}, Gemini-3.1 Pro~\cite{googledeepming2026gemini31pro}\footnote{Proprietary models are used in their chat version.}, along with human-generated ones. These are passed to \texttt{openai/gpt-oss-20b} and \texttt{google/gemma-3-4b-it} in order to be guessed and the results are reported in Figure~\ref{fig:models-vs-models}. \texttt{openai/gpt-oss-20b} assigns near-zero probability to the correct target across virtually all clue givers, reaching at most 1.2\% at pass@10 on human descriptions. \texttt{google/gemma-3-4b-it} is substantially more capable, yet still requires between 5 and 10 guesses to approach a decent accuracy: on human-written descriptions, its pass@1 is 0.0\% while pass@10 reaches 30.0\%. The same asymmetry holds when models guess from model-written descriptions, with Claude Sonnet-4.6 and Gemini-3.1-Pro as clue givers yielding comparable pass@10 of 25.0\% and 30.0\% respectively under \texttt{google/gemma-3-4b-it} judge, while \texttt{openai/gpt-oss-20b} remains at or below 20.0\% across all clue givers.


\subsection{Humans versus models; models versus humans}


\paragraph{Humans as guessers.}
We asked human annotators to read model-generated descriptions and guess the target word. Overall, annotators correctly identified the target in 30.77\% of cases (weighted average across annotators and conditions), with performance varying 
considerably by constraining strategy: constrained generation yields the highest human accuracy (38.10\%), followed by the SAE-based condition (31.03\%), with prompting producing the lowest (23.44\%). This ordering mirrors the pattern observed in Section~\ref{sec:model_guess} for model judges, suggesting that the informativeness advantage of constrained generation generalises beyond automatic evaluation.

Beyond aggregate accuracy, qualitative inspection reveals recurring failure patterns in model-generated descriptions, including referential failures, category mismatches, and code-switching under the SAE condition, which we discuss in 
Appendix~\ref{app:failure-patterns}.

\paragraph{Humans as descriptors.}


We also collected human-generated descriptions across three conditions that differ in prior exposure and time constraints (see \textit{description phase} in Section~\ref{sec:evaluation}): \textit{spontaneous} (no preparation, 5~cards), \textit{deliberate} (time to reflect before writing, 5~cards), and \textit{shared} (all annotators describe privately the same ten words, enabling direct comparison with models). For the \textit{spontaneous} and \textit{deliberate} conditions, we assessed how accurately other annotators could guess the target word from the description: \textit{deliberate} descriptions achieve 85\% accuracy, while \textit{spontaneous} descriptions yield a higher rate of 95\%. The gap between these two reflects the interactive nature of oral description, where players could refine in real time, versus the deliberate setting, which more closely mirrors the constraints imposed on models. 

If we ask the models to guess the human-described words, we can spot a clear ordering: shared $>$ deliberate $>$ spontaneous at all values of $k$ (Table~\ref{tab:human_eval_full}). At pass@10, shared descriptions yield 15.6\%, deliberate 13.8\%, and spontaneous only 5.1\%. As observed throughout, \texttt{openai/gpt-oss-20b} almost never guesses correctly (pass@10 $\leq$ 2.5\%), whereas \texttt{google/gemma-3-4b-it} reaches up to 30.0\% on shared descriptions, yet still requiring between 5 and 10 guesses to rival what a human achieves at pass@1. 


This contrast is revealing: humans benefit from spontaneous descriptions that, though fragmented, are iteratively refined towards the referent in real time. When presented to models as static text (without the interactive context that makes them effective for human guessers) such descriptions yield substantially lower pass@$k$ rates, while models are better suited to process the structured, fixed outputs of the deliberate and shared conditions.

\begin{table}[ht]
\vspace{-6pt}
\centering
\small
\caption{Human clue-giver evaluation. Accuracy computed on strict criterion (no exact + no morphological violations).}
\label{tab:human_eval_full}
\begin{tabular}{llccccccc}
\toprule
\multirow{2}{*}{\textbf{Condition}} & \multirow{2}{*}{\textbf{Judge}} & \multicolumn{5}{c}{\textbf{pass@$k$}} & \multicolumn{2}{c}{\textbf{Accuracy}} \\
\cmidrule(lr){3-7} \cmidrule(lr){8-9}
 & & @1 & @2 & @3 & @5 & @10 & loose & strict \\
\midrule
\multirow{2}{*}{\parbox{2cm}{Spontaneous\\($n=39$)}}
  & \texttt{gpt-oss-20b}   & 0.0\% & 0.0\% & 0.0\% & 0.0\% &  0.0\% & \multirow{2}{*}{97.4\%} & \multirow{2}{*}{82.1\%} \\[2pt]
  & \texttt{gemma-3-4b-it} & 0.0\% & 2.6\% & 5.1\% & 7.7\% & 10.3\% & & \\
\midrule
\multirow{2}{*}{\parbox{2cm}{Deliberate\\($n=40$)}}
  & \texttt{gpt-oss-20b}   & 0.0\% & 0.0\% & 0.0\% &  0.0\% &  2.5\% & \multirow{2}{*}{100.0\%} & \multirow{2}{*}{97.5\%} \\[2pt]
  & \texttt{gemma-3-4b-it} & 0.0\% & 5.0\% & 7.5\% & 10.0\% & 25.0\% & & \\
\midrule
\multirow{2}{*}{\parbox{2cm}{Shared\\($n=80$)}}
  & \texttt{gpt-oss-20b}   & 0.0\% & 0.0\% &  0.0\% &  1.2\% &  1.2\% & \multirow{2}{*}{100.0\%} & \multirow{2}{*}{91.2\%} \\[2pt]
  & \texttt{gemma-3-4b-it} & 0.0\% & 7.5\% & 12.5\% & 18.8\% & 30.0\% & & \\
\bottomrule
\end{tabular}
\vspace{-8pt}
\end{table}

\section{Discussion}

Our results reveal a consistent tension between constraint adherence and descriptive informativeness that cuts across all conditions. The three families of constraining strategies we examine operate at qualitatively different levels (instruction, decoding, and internal representation) and this difference determines not only how reliably the model avoids forbidden words, but also how it describes the target item. Prompting achieves high compliance at the cost of informationally sparse output; constrained generation forces the model toward more semantically precise paraphrases, at the cost of occasionally leaking the target word; SAE-based steering sits at the opposite extreme, preserving conceptual associations at the cost of surface compliance. This trade-off suggests that lexical avoidance and communicative effectiveness are not independently controllable, and that the mechanism through which a constraint is imposed shapes the output in ways that go beyond mere compliance.

A second thread concerns the asymmetry between humans and models as players. On the descriptor side, reasoning does help models: \texttt{openai/gpt-oss-20b} with chain-of-thought reaches 97.4\% accuracy versus 72.2\% without, a substantial gain. For humans, the picture is more nuanced: spontaneous oral descriptions, despite being unplanned, yield higher guessing accuracy (95.0\%) than deliberate written ones (85.0\%), likely because the interactive oral setting allows real-time refinement. This suggests that planning confers different benefits depending on the medium and the player: for models it primarily enforces compliance, whereas for humans it is the interactivity of the setting, rather than deliberation \textit{per se}, that drives performance. On the guesser side, the asymmetry is starker: both judge models struggle to identify target words, with \texttt{google/gemma-3-4b-it} requiring between 5 and 10 guesses to approach the accuracy a human achieves at pass@1. This gap suggests that models do not share the same network of salient lexical associations that makes Taboo intuitive for humans, and that guessing from indirect descriptions remains a genuinely hard task for current language models. This gap is also reminiscent of the distinction between fast, associative System~1 processing and deliberate System~2 reasoning \citep{kahneman2011thinking}: humans navigate Taboo by rapidly activating and suppressing salient lexical associates, a mode of processing that current language models do not appear to replicate, at least not in the guessing direction.

Our study has however several limitations. The benchmark is currently Italian-only, and it is unclear how results generalise to languages with different morphological profiles or lexical structures. The shared evaluation set covers only 10 words, limiting the statistical power of cross-model comparisons in Section~\ref{sec:model_guess}. SAE-based steering was evaluated on \texttt{google/gemma-3-4b-it} only, and extending it to other architectures remains open. The two judge models differ substantially in sensitivity, raising questions about which better reflects human guessing behaviour, a question our human evaluation begins to address but does not fully resolve.

\section{Conclusions and Future Work}

We presented a framework that operationalises lexical constraint following as a communicative game, and used it to evaluate three families of constraining strategies across multiple models and human annotators. Our results show that compliance and communicative effectiveness are not independently controllable: the mechanism through which a constraint is imposed (instruction, decoding, or internal representation) shapes the output in ways that go beyond mere rule adherence, with constrained generation and SAE-based steering producing qualitatively different trade-offs. Beyond the specific findings, our work supports the broader argument that games constitute a natural and underexplored testbed for language model capabilities: they impose well-defined rules, require communicative grounding, and admit quantitative evaluation while remaining ecologically valid. LMtaboo in particular is inherently interactive, yet we have evaluated models only in a single-turn, single-agent setting. Extending the benchmark to multi-turn and multi-agent configurations (where models iteratively refine descriptions based on guesser feedback, or negotiate meaning across turns) would more faithfully reflect the dynamics of the original game and probe capabilities, such as pragmatic adaptation and collaborative grounding, that current evaluations leave largely untested.

\clearpage

\begin{acknowledgments}
The authors sincerely thank Valentina, Frida, Daniel, and Arianna for their invaluable contribution to the annotation process.
The work of Daniel Scalena has been partially funded by MUR under the grant ReGAInS, \textit{Dipartimenti di Eccellenza 2023-2027} of the Department of Informatics, Systems and Communication at the University of Milano-Bicocca.
The work of Sara Candussio has been funded by Fondo Sociale Europeo Plus of Regione Autonoma Friuli Venezia Giulia. 
We also thank the Center for Information Technology of the University of Groningen for providing access to the H\'abr\'ok high-performance computing cluster used for part of the experiments.

\end{acknowledgments}

\section*{Declaration on Generative AI}
During the preparation of this work, the authors used \textbf{Claude} (Anthropic) in order to \textbf{improve writing style}, \textbf{grammar and spelling check}, \textbf{paraphrase and reword}, \textbf{drafting content} (including experimental code). After using this tool/service, the authors reviewed and edited the content as needed and take full responsibility for the publication's content.

\bibliography{sample-ceur}

\appendix

\clearpage

\section{Appendix A: prompt details, generation parameters and repetition penalty}\label{app:prompts}

This appendix reports the prompt templates used in our experiments, together with the generation hyperparameters adopted for each model and condition.

\subsection{Prompt Templates}

Two prompt variants were used. The \textbf{baseline} prompt asks the model to describe the target word freely, with no mention of forbidden words. The \textbf{prompted} condition additionally instructs the model to avoid the forbidden words and their morphological relatives.

\begin{table}[h]
\centering
\footnotesize
\begin{tabular}{p{0.9\columnwidth}}
\hline\\[-4pt]
\textbf{User Message — Baseline} \\[2pt]
\hline\\[-4pt]
Descrivi la seguente parola in modo che qualcuno possa indovinarla. \\[2pt]
Parola da descrivere: \texttt{[target word]} \\[2pt]
Scrivi in output solamente la descrizione senza nessun testo di contorno. Considera che hai un limite di 30 parole, quindi cerca di dare una descrizione quanto più breve possibile. \\[2pt]
Descrizione: \\[2pt]
\hline\\[-4pt]
\textit{Describe the following word so that someone can guess it.}\\[2pt]
\textit{Word to describe: \texttt{[target word]}}\\[2pt]
\textit{Output only the description with no surrounding text. You have a limit of 30 words, so try to give as brief a description as possible.}\\[2pt]
\textit{Description:}\\[2pt]
\hline\\[2pt]
\textbf{User Message — Prompted condition} \\[2pt]
\hline\\[-4pt]
Descrivi la seguente parola in modo che qualcuno possa indovinarla, ma NON usare nessuna di queste parole o nessuna parola con la stessa radice: \texttt{[fw\textsubscript{1}, fw\textsubscript{2}, \ldots]}. \\[2pt]
Parola da descrivere: \texttt{[target word]} \\[2pt]
Scrivi in output solamente la descrizione senza nessun testo di contorno. Considera che hai un limite di 30 parole, quindi cerca di dare una descrizione quanto più breve possibile. \\[2pt]
Descrizione: \\[2pt]
\hline\\[-4pt]
\textit{Describe the following word so that someone can guess it, but do NOT use any of these words or any word with the same root: \texttt{[fw\textsubscript{1}, fw\textsubscript{2}, \ldots]}.}\\[2pt]
\textit{Word to describe: \texttt{[target word]}}\\[2pt]
\textit{Output only the description with no surrounding text. You have a limit of 30 words, so try to give as brief a description as possible.}\\[2pt]
\textit{Description:}\\[2pt]
\hline
\end{tabular}
\caption{Prompt templates for the \textbf{baseline} (top) and \textbf{prompted} (bottom) conditions. English translation in italics.}
\label{tab:prompts}
\end{table}

\subsection{Generation parameters and repetition penalty}

All models were run in greedy decoding mode (\texttt{do\_sample=False}). Table~\ref{tab:gen_params}summarises the generation parameters used per model and condition.

\begin{table}[h]
\centering
\small
\begin{tabular}{lcc}
\toprule
\textbf{Model} & \textbf{max\_new\_tokens} & \textbf{rep.\ penalty} \\
\midrule
\texttt{gemma-3-4b-it}  & 80   & 1.0 \\
\texttt{gpt-oss-20b} (reasoning)    & 2048 & 1.2 \\
\texttt{gpt-oss-20b} (no reasoning) & 80   & 1.0 \\
\bottomrule
\end{tabular}
\caption{Generation hyperparameters per model and condition.}
\label{tab:gen_params}
\end{table}

For \texttt{openai/gpt-oss-20b} in the reasoning condition, the assistant turn was left open so the model could produce a chain-of-thought trace before committing to a final answer. In the no-reasoning condition, the assistant turn was instead prefilled with a channel token that routes the model directly to its final-answer channel, bypassing the reasoning trace. A repetition penalty of $1.2$ was applied exclusively in the reasoning condition to mitigate degenerate repetition loops that can arise during long generations; no penalty was applied otherwise.

\section{Appendix B: a priori constrained generation vs online}\label{app:a_priori}

Both the \textit{a priori} and the inference-time (online) approaches enforce the taboo constraint by operating on stems and lemmas of the forbidden words, using the same morphological pipeline (spaCy \texttt{it\_core\_news\_sm} + \texttt{nltk} Italian Stemmer). They differ fundamentally in \emph{when} and \emph{how broadly} the constraint is applied.

\paragraph{A priori censorship}
Before generation begins, the method computes a static set of banned token IDs: for each forbidden word, all stems and lemmas are extracted, and every vocabulary token whose surface form shares a prefix with any of them (in either direction) is masked by forcing its logit to $-\infty$ for the entire generation.
This static mask is context-free: a token is banned regardless of whether its occurrence in the output would actually constitute a violation.

The consequences can be severe. Consider the target word \textit{accipicchia} (an Italian exclamation of surprise), whose forbidden list includes \textit{per la miseria} (a common interjection meaning "what a misery"). Because the phrase contains the function words \textit{per} ("for") and \textit{la} ("the"), the stemmer extracts them as forbidden stems, causing \num{2171} vocabulary tokens to be masked --- including common prepositions, determiners, and unrelated content words that merely share a surface prefix with \textit{per} or \textit{la} (e.g.\ tokens such as \texttt{\_permitting}, \texttt{\_lancé}, \texttt{larghezza}).
Across the full dataset of 194 items, the \textit{a priori} approach bans a cumulative total of \num{90957} tokens (on average \num{469} per item, out of a vocabulary of \num{262145}).

\paragraph{Inference-time constrained generation}
The online approach bans \emph{no tokens} before generation begins. Instead, it generates freely token by token, accumulates tokens into complete words, and only when a word boundary is detected checks whether the completed word matches any forbidden stem or lemma. If a match is found, the method backtracks to the first token of the offending word and re-samples, blocking only that specific token at that position.
For \textit{accipicchia}, the online approach evaluates each generated word against the forbidden forms at runtime, and only intervenes if a word such as \textit{meraviglia} or \textit{caspita} is actually produced --- leaving \textit{per}, \textit{la}, and all other innocent tokens freely available.
Across the same 194 items, the model produces on average \num{19.3} words per output, all of which are checked at runtime --- yet zero violations are recorded, confirming that the online constraint intervenes surgically only when needed rather than preemptively suppressing large portions of the vocabulary.

\section{Appendix C: SAE implementation details}
\label{app:sae-impl}

\paragraph{SAE configuration}
We use JumpReLU SAE from \texttt{google/gemma-scope-2-4b-it}~\citep{lieberum-etal-2024-gemma}, applied to the residual stream of \texttt{google/gemma-3-4b-it}, which has 34 transformers layers in its text module. The Gemma Scope 2 checkpoints provides SAEs at four depths: layer 9, 17, 22 and 29 (approximately $25\%$, $50\%$, $65\%$ and $85\%$ of network depth). We select layer 29, as later layers are known to carry richer semantic representation compared to early layers~\citep{skean2025layer}, making them better suited for identifying concept-level features associated with specific words.
The SAE uses a dictionary width of 65k features and medium L0 sparsity, with weights loaded in \texttt{bfloat16} and kept frozen throughout. Feature identification runs once per forbidden word per card and is cached to avoid redundant forward passes across items sharing the same forbidden word. The feature prompt template \textit{"La parola vietata è: \{word\}"} is tokenized with special tokens; the token span of $w_i$ is located by matching the tokenized word (with and without a leading space) against the full input sequence, falling back to the final token if no march is found.

\section{Appendix E: Failure Patterns in Model-Generated Descriptions}
\label{app:failure-patterns}

Beyond aggregate accuracy, qualitative inspection of the guessing phase reveals several recurring failure patterns in model-generated descriptions.

\paragraph{Code-switching} 
While descriptions generated under the prompting condition remain consistently in Italian, constrained generation introduces a small but noticeable rate of language shifts (6.35\%), which becomes far more pronounced under the SAE condition (44.83\%), with English, French, and Spanish being the most frequent alternates. This suggests that intervening on internal representations disrupts the model's linguistic behaviour in ways that explicit lexical guidance does not.

\paragraph{Referential failure} 
A significant proportion of descriptions are nonsensical or semantically detached from the target word, making guessing effectively impossible. Out of 160 ground-truth entries evaluated, 25 (15.62\%) were flagged as containing an inappropriate description. A striking example is the word \textit{semolino} (semolina), described under two conditions by the same model as follows:

\begin{itemize}
    \item \texttt{gemma\_sae}: \textit{``Piccola protuberanza dura e rotonda, spesso 
    presente sulla pelle, solitamente causata da irritazione o crescita di 
    foruncoli.''} (\textit{``A small, hard, round bump, often found on the skin, 
    usually caused by irritation or the growth of pimples.''})
    \item \texttt{gemma\_constrained}: \textit{``Piccola goccia di liquido, spesso 
    salata, che si stacca da una fonte, come una lacrima o una pioggia leggera.''} 
    (\textit{``A small drop of liquid, often salty, that detaches from a source, 
    such as a tear or light rain.''})
\end{itemize}

Neither description bears any semantic relation to the target word: the generated text is fluent and well-formed, but lacks any associative link to the intended referent.

\paragraph{Category mismatch} 
In several cases, the model describes a semantically 
related but categorically distinct concept --- for instance, foregrounding the activity rather than the agent, or the domain rather than the practitioner. An example is the word \textit{falegname} (carpenter), described by \texttt{gpt\_think\_constrained} as:

\begin{itemize}
    \item \texttt{gpt\_think\_constrained}: \textit{``Arte che trasforma il legno: 
    taglia, intaglia, costruisce mobili e strutture con seghe, scalpelli e 
    martelli.''} (\textit{``The art of transforming wood: cutting, carving, building 
    furniture and structures with saws, chisels and hammers.''})
\end{itemize}

By framing the description around the craft (\textit{arte}) rather than the practitioner, the model effectively steers the guesser away from the intended referent.

\clearpage

\section{Tables}
\label{app:results}

\noindent\textbf{Note on evaluation metrics.} The metrics reported in the Eval section of each table are independent counters and do not sum to 100\%. A single output may simultaneously trigger multiple error categories (e.g., a target leak and a filter failure), so the categories are not mutually exclusive.

\begin{table}[h]
\centering
\small
\setlength{\tabcolsep}{5pt}
\renewcommand{\arraystretch}{1.15}
\begin{tabular}{l l ccccc}
\toprule
& &
{Prompting} &
\multicolumn{2}{c}{Constrained} &
{Manip.} & {Baseline} \\
\cmidrule(lr){3-3} \cmidrule(lr){4-5} \cmidrule(lr){6-6} \cmidrule(lr){7-7}
\textbf{Section} & \textbf{Metric} &
{} & {a priori} & {online} & {SAEs} & {no forb.} \\
\midrule
\multirow{5}{*}{\textbf{Eval}}
& True accuracy             & 91.20 & 98.50 & 96.40 & 55.70 & 34.00 \\
& \% exact viol.            &  2.60 &  0.00 &  0.00 & 31.40 & 52.10 \\
& \% only morphol.\ viol.   &  4.60 &  0.00 &  0.00 & 11.90 & 12.30 \\
& \% no \texttt{<eot>}      &  0.00 &  0.00 &  0.00 &  0.00 &  0.00 \\
& \% target leaked          &  1.50 &  1.50 &  3.60 &  1.00 &  1.50 \\
\midrule
\multirow{11}{*}{\textbf{gpt-oss judge}}
& pass@1  & 0.00 & 0.00 & 0.00 & 0.00 & 0.00 \\
& pass@2  & 0.00 & 0.50 & 0.50 & 0.00 & 0.50 \\
& pass@3  & 1.00 & 1.50 & 2.60 & 1.00 & 3.60 \\
& pass@5  & 1.50 & 3.60 & 5.20 & 3.10 & 5.70 \\
& pass@10 & 3.60 & 7.70 & 8.80 & 5.70 & 9.30 \\
\cmidrule(lr){2-7}
& min    & 4e-6   & 4e-6   & 4e-6   & 1e-6   & 4e-6   \\
& p25    & 1.1e-5 & 3.0e-5 & 2.6e-5 & 1.0e-5 & 2.7e-5 \\
& median & 1.9e-5 & 6.9e-5 & 6.2e-5 & 1.9e-5 & 1.2e-4 \\
& avg    & 1.9e-3 & 3.4e-3 & 4.5e-3 & 1.8e-3 & 5.0e-3 \\
& p75    & 1.3e-4 & 3.5e-4 & 6.1e-4 & 1.6e-4 & 1.3e-3 \\
& max    & 0.1143 & 0.1564 & 0.1765 & 0.0576 & 0.1437 \\
\midrule
\multirow{11}{*}{\textbf{gemma judge}}
& pass@1  &  0.00 &  0.00 &  0.00 &  0.00 &  0.00 \\
& pass@2  &  1.50 &  2.10 &  4.60 &  4.10 &  3.10 \\
& pass@3  &  4.10 &  4.60 &  7.70 &  7.20 &  7.20 \\
& pass@5  &  6.20 &  9.80 &  9.80 & 13.90 & 11.90 \\
& pass@10 & 11.30 & 17.50 & 17.00 & 21.10 & 24.70 \\
\cmidrule(lr){2-7}
& min    & 0      & 0      & 0      & 0      & 0      \\
& p25    & 0      & 0      & 0      & 0      & 0      \\
& median & 0      & 0      & 0      & 0      & 0      \\
& avg    & 6.9e-4 & 0      & 1.4e-4 & 4.5e-4 & 1.9e-4 \\
& p75    & 4e-6   & 3e-6   & 2e-6   & 1.2e-5 & 2e-6   \\
& max    & 0.1272 & 3.2e-3 & 1.8e-2 & 4.7e-2 & 1.8e-2 \\
\bottomrule
\end{tabular}
\caption{Evaluation results -- \textbf{gemma-3-4b-it}}
\label{tab:results-gemma}
\end{table}

\begin{table}[h]
\centering
\small
\setlength{\tabcolsep}{5pt}
\renewcommand{\arraystretch}{1.15}
\begin{tabular}{l l cccc}
\toprule
& &
{Prompting} &
\multicolumn{2}{c}{Constrained} &
{Baseline} \\
\cmidrule(lr){4-5} \cmidrule(lr){6-6}
\textbf{Section} & \textbf{Metric} &
{} & {a priori} & {online} & {no forb.} \\
\midrule
\multirow{5}{*}{\textbf{Eval}}
& True accuracy             & 72.20 & 82.50 & 82.00 & 24.70 \\
& \% exact viol.            & 16.00 &  0.00 &  0.00 & 55.20 \\
& \% only morphol.\ viol.   &  8.70 &  0.00 &  0.50 & 14.90 \\
& \% no \texttt{<eot>}      &  0.00 &  0.00 &  0.00 &  0.00 \\
& \% target leaked          &  3.10 & 17.50 & 17.50 & 12.90 \\
\midrule
\multirow{11}{*}{\textbf{gpt-oss judge}}
& pass@1  & 0.00 & 0.00 & 0.00 & 0.00 \\
& pass@2  & 0.00 & 0.00 & 0.00 & 0.00 \\
& pass@3  & 0.50 & 0.50 & 0.50 & 1.00 \\
& pass@5  & 1.00 & 0.50 & 0.50 & 2.60 \\
& pass@10 & 3.60 & 2.10 & 2.10 & 6.20 \\
\cmidrule(lr){2-6}
& min    & 2e-6   & 2e-6   & 1e-6   & 1e-6   \\
& p25    & 1.1e-5 & 1.1e-5 & 1.1e-5 & 1.2e-5 \\
& median & 2.5e-5 & 1.6e-5 & 1.5e-5 & 3.2e-5 \\
& avg    & 8.6e-4 & 8.7e-5 & 9.0e-5 & 1.8e-3 \\
& p75    & 1.4e-4 & 3.0e-5 & 2.6e-5 & 3.9e-4 \\
& max    & 0.0635 & 3.3e-3 & 5.6e-3 & 0.1416 \\
\midrule
\multirow{11}{*}{\textbf{gemma judge}}
& pass@1  &  0.00 &  0.00 &  0.00 &  0.00 \\
& pass@2  &  2.10 &  6.70 &  6.70 &  3.10 \\
& pass@3  &  4.10 & 11.30 & 13.90 &  7.70 \\
& pass@5  & 10.30 & 17.50 & 19.10 & 11.30 \\
& pass@10 & 18.60 & 24.20 & 24.70 & 17.50 \\
\cmidrule(lr){2-6}
& min    & 0      & 0      & 0      & 0      \\
& p25    & 0      & 0      & 0      & 0      \\
& median & 0      & 0      & 0      & 0      \\
& avg    & 4.4e-4 & 2.8e-3 & 1.1e-3 & 1.1e-3 \\
& p75    & 4e-6   & 3.7e-5 & 4.3e-5 & 4e-6   \\
& max    & 0.0366 & 0.2456 & 0.0498 & 0.1358 \\
\bottomrule
\end{tabular}
\caption{Evaluation results -- \textbf{gpt-oss-20b (no reasoning)}}
\label{tab:results-gptoss-nreas}
\end{table}

\begin{table}[t]
\centering
\small
\setlength{\tabcolsep}{5pt}
\renewcommand{\arraystretch}{1.15}
\begin{tabular}{l l cccc}
\toprule
& &
{Prompting} &
\multicolumn{2}{c}{Constrained} &
{Baseline} \\
\cmidrule(lr){4-5} \cmidrule(lr){6-6}
\textbf{Section} & \textbf{Metric} &
{} & {a priori} & {online} & {no forb.} \\
\midrule
\multirow{5}{*}{\textbf{Eval}}
& True accuracy             & 97.40 & 92.30 & 87.10 & 34.00 \\
& \% exact viol.            &  1.00 &  0.00 &  1.00 & 45.90 \\
& \% only morphol.\ viol.   &  0.50 &  0.00 &  0.50 & 17.50 \\
& \% no \texttt{<eot>}      &  1.00 &  2.10 &  2.10 &  0.00 \\
& \% target leaked          &  2.10 &  7.70 & 12.40 &  4.10 \\
\midrule
\multirow{11}{*}{\textbf{gpt-oss judge}}
& pass@1  & 0.00 & 0.50 & 0.00 & 0.00 \\
& pass@2  & 0.00 & 1.00 & 0.50 & 0.50 \\
& pass@3  & 0.00 & 1.50 & 0.50 & 0.50 \\
& pass@5  & 0.50 & 2.10 & 1.00 & 4.60 \\
& pass@10 & 1.50 & 5.70 & 4.60 & 6.70 \\
\cmidrule(lr){2-6}
& min    & 1e-6   & 2e-6   & 2e-6   & 1e-6   \\
& p25    & 1.0e-5 & 1.1e-5 & 1.0e-5 & 1.4e-5 \\
& median & 1.8e-5 & 2.7e-5 & 2.2e-5 & 3.8e-5 \\
& avg    & 3.4e-4 & 3.5e-3 & 1.9e-3 & 2.3e-3 \\
& p75    & 6.1e-5 & 1.6e-4 & 1.9e-4 & 2.5e-4 \\
& max    & 0.0269 & 0.2779 & 0.1916 & 0.1281 \\
\midrule
\multirow{11}{*}{\textbf{gemma judge}}
& pass@1  &  0.50 &  0.00 &  0.00 &  0.00 \\
& pass@2  &  4.60 &  5.20 &  4.10 &  4.10 \\
& pass@3  &  6.70 &  7.70 &  7.70 &  9.80 \\
& pass@5  & 11.30 & 13.90 & 10.30 & 17.50 \\
& pass@10 & 16.50 & 19.60 & 20.60 & 21.60 \\
\cmidrule(lr){2-6}
& min    & 0      & 0      & 0      & 0      \\
& p25    & 0      & 0      & 0      & 0      \\
& median & 0      & 0      & 0      & 0      \\
& avg    & 2.2e-3 & 7.0e-4 & 3.6e-4 & 4.7e-4 \\
& p75    & 5e-6   & 8e-6   & 8e-6   & 9e-6   \\
& max    & 0.2870 & 0.0746 & 0.0228 & 0.0309 \\
\bottomrule
\end{tabular}
\caption{Evaluation results -- \textbf{gpt-oss-20b}}
\label{tab:results-gptoss-repen}
\end{table}

\end{document}